# Accurate and Reliable Methods for 5G UAV Jamming Identification With Calibrated Uncertainty

Hamed Farkhari∗†§, Joseanne Viana∗‡§, Pedro Sebastião ∗‡, Luis Miguel Campos †, Luis Bernardo ‡, Rui Dinis‡, Sarang Kahvazadeh ¶

∗ISCTE – Instituto Universitário de Lisboa, Av. das Forças Armadas, 1649-026 Lisbon, Portugal

†PDMFC, Rua Fradesso da Silveira, n. 4, Piso 1B, 1300-609, Lisboa, Portugal

‡IT – Instituto de Telecomunicações, Av. Rovisco Pais, 1, Torre Norte, Piso 10, 1049-001 Lisboa, Portugal

FCT – Universidade Nova de Lisboa, Monte da Caparica, 2829-516 Caparica, Portugal;

¶ CTTC - Centre Tecnològic de Telecomunicacions de Catalunya (CERCA);

Emails : {Hamed_Farkhari, joseanne_cristina_viana}@iscte-iul.pt

*Abstract*—Only increasing *accuracy* without considering uncertainty may negatively impact Deep Neural Network (DNN) decision-making and decrease its reliability. This paper proposes five combined preprocessing and post-processing methods for time-series binary classification problems that simultaneously increase the *accuracy* and reliability of DNN outputs applied in a 5G UAV security dataset. These techniques use DNN outputs as input parameters and process them in different ways. Two methods use a well-known Machine Learning (ML) algorithm as a complement, and the other three use only confidence values that the DNN estimates. We compare seven different metrics, such as the Expected Calibration Error (ECE), Maximum Calibration Error (MCE), Mean Confidence (MC), Mean Accuracy (MA), Normalized Negative Log Likelihood (NLL), Brier Score Loss (BSL), and Reliability Score (RS) and the tradeoffs between them to evaluate the proposed hybrid algorithms. First, we show that the eXtreme Gradient Boosting (XGB) classifier might not be reliable for binary classification under the conditions this work presents. Second, we demonstrate that at least one of the potential methods can achieve better results than the classification in the DNN softmax layer. Finally, we show that the prospective methods may improve *accuracy* and reliability with better uncertainty calibration based on the assumption that the RS determines the difference between MC and MA metrics, and this difference should be zero to increase reliability. For example, Method 3 presents the best RS of 0.65 even when compared to the XGB classifier, which achieves RS of 7.22.

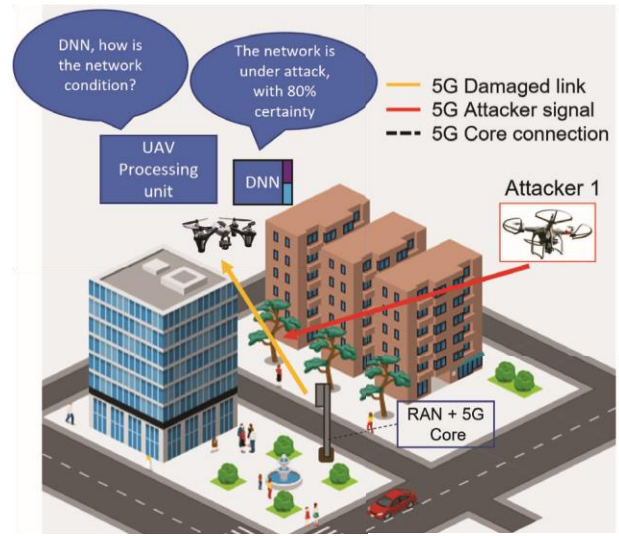

Fig. 1: Proposed scenario and example of inherited uncertainty in the deep network outputs when applied in UAV security field.

## I. INTRODUCTION

Deep Neural Networks (DNNs) have seen extensive deployment due to their recent achievements in several fields. Prediction distributions generated by such models increasingly make decisions in the telecommunications and security sectors [1, 2, 3].

For example, 6G telecommunication systems will incorporate Machine Learning (ML) mechanisms such as DNNs into their standards [1] and there are several studies on how to apply deep learning decision-making in the physical layer [2]. Another promising field for DNN applications is 5G UAV security [4, 5]. DNNs are interesting to use due to their universal function capabilities, superior logic that allows them to solve complex time series modeling issues, and depending on their design, the possibility to process data in parallel. However, due to the DNN's iterative data processing, classification applications can provide probabilities with uncertainties in the outputs which raise concerns about the reliability of the true correctness likelihood of its classification decisions. Fig. 1 presents an example of the inherited uncertainty in the DNN outputs applied to UAV wireless communication security. Such estimates describe the conditions under which the results of the model can be trusted or not.

The authors in [6] discuss the importance of calibrating DNNs in order to guarantee high *accuracy* and reliable output decisions. They show at least six calibration techniques that increase both parameters in widely recognized datasets (i.e., CIFAR-10 and ImageNet) applied in pretrained DNNs (i.e., RestNet, WideNet, and LeNet). In [7], the authors justify the need to specify the uncertainty especially in critical realworld settings, in which the input distribution deviates from the training distribution because of sample bias and nonstationarity.

§Collaborative authors with equal contribution.

Understanding questions of risk, uncertainty, and trust in a model's output becomes increasingly important when augmented techniques are used at the original data

## II. SYSTEM MODEL

Fig. 2 depicts the block diagram of the proposed system. On the left side, we have the proposed processing algorithm. In the preprocessing stage. The authors in [8] suggest that prepossessing and postapplied before and after DNN processing, respectively.

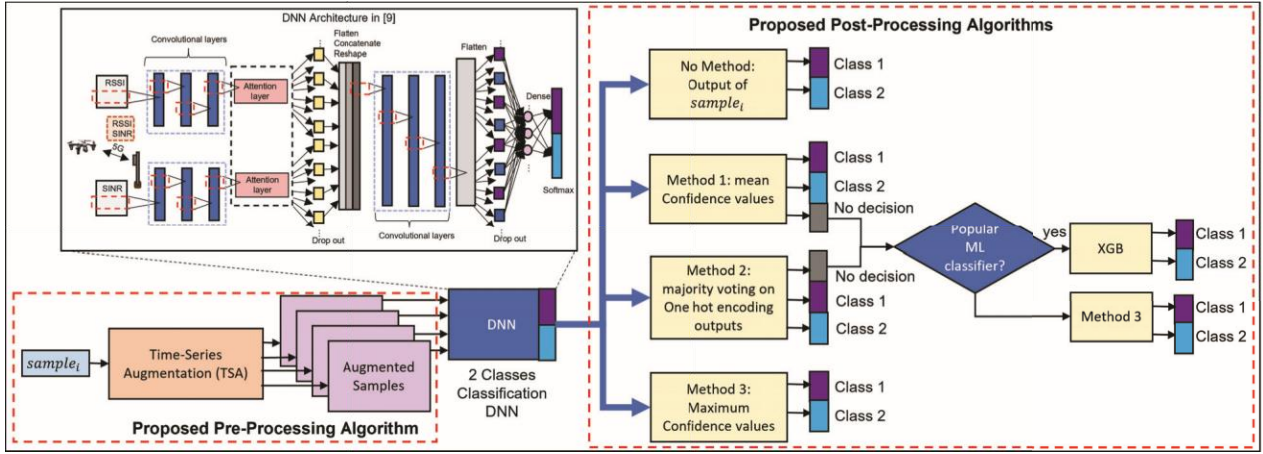

Fig. 2: Highlighted proposed combined preprocessing and post-processing algorithms (No Method, and Methods 1, 2 and 3)

processing techniques can improve $N$ inputs and $M$ class DNNs. The authors in [6, 7] also propose methods that increase *accuracy* while reducing uncertainty in classification tasks and mathematical approaches to calculate the Expected Calibration Error (ECE), the Maximum Calibration Error (MCE), and estimate if the DNN is over-confident or underconfident.

Inspired by the possibility of choosing a tolerable degree of uncertainty and increasing the reliability of DNN outputs used in 5G UAV security, this study presents five new combined prepossessing and post-processing techniques that increase the overall *accuracy* and reliability of binary classification deep networks by adjusting the uncertainty. We assess these methods using seven key performance metrics related to errors in calibration and in confidence values. Then, we utilize the Reliability Score (RS) that measures the difference between the Mean Accuracy (MA) and Mean Confidence (MC) to measure the degree of uncertainty. Finally, we evaluate the proposed algorithms' impact on the DNN's performance compared to the baseline DNN with no algorithms applied and the DNN added to the eXtreme Gradient Boosting (XGB) classifier. The XGB classifier is selected because of its superior *accuracy* in comparison to five other classifiers we test with our data [9].

The paper is structured as follows: Section II describes all the components of the experiment. Subsection II-A illustrates the small deep network used. The dataset is explained in subsection II-B. Subsection II-C details the five combined preprocessing and post-processing techniques and how they might increase reliability and *accuracy* at the same time. Subsection II-D presents the metrics to evaluate reliability for each method. At the end, section III shows the simulation results for the prospective system. Lastly, section IV describes the main conclusions.

middle, we have the DNN presented by [5] and on the right side, we present the components of the thee proposed post-processing algorithms. The preprocessing Algorithm contains three blocks: the input $sample_i$, the TimeSeries Augmentation technique, and the Augmented samples. The Post-Processing Algorithm includes Methods 1, 2, and 3 and their possible complements named XGB Algorithm and Method 3, respectively. We add the "No Method" block in the system to be able to compare the improvements related to the *accuracy* and reliability of each algorithm with the features classification results available at the end of the DNN processing without any method applied. We choose to apply the XGB classifier algorithm to classify features located in the "No decision" class for Method 1 + XGB and Method 2 + XGB because it performs the best when classifying either the presence or absence of attacks for different scenarios and configurations using the dataset in II-B as discussed in [9, 10]. Any other high performing ML algorithm can be used. Note that the XGB algorithm is used as a complement to the proposed Methods 1 and 2. We use only the original sample as an input for the XGB algorithm and we propose to replace it by Method 3.

### A. Deep Network architecture

We analyse the confidence values in the deep network proposed by [9, 10] in terms of reliability. The simplified multi headed DNN Architecture illustrated in Fig. 2 contain the following layers: (i) three Convolutional layers, (ii) an Attention layer, and (iii) a Drop out layer in each head. The body of the deep network consists of: (i) a Flatten, (ii) a Concatenate, (iii) a Reshape, (iv) three Convolutional layers, (v) a Flatten, (vi) a Drop out layer, (vii) a Fully connected layer, and (viii) the output layer for two classes classification.

### B. Dataset

Our dataset contains data from the Received Signal Strength Indicator (RSSI) and the Signal to Interference-plus-Noise Ratio (SINR) measurements collected when an authenticated

UAV is connected in the small cell through the 5G communication system, and there are power attacks from other UAVs in the network. There are other terrestrial users connected to the network. The measured parameters in the authenticated UAV change as the interference from the other devices increases or decreases. More details on the dataset construction and one possible application for the dataset is available in [10] and in [11]. Fig. 1 presents the proposed scenario and illustrates only one attacker and twelve terrestrial users, but the dataset covers up to four attackers and 30 terrestrial users connected simultaneously. We are also studying and analyzing other open-source datasets, such as WSN-DS [12], and [13] to have a diverse dataset with mature data.

### C. Designed Solution

In this section, we explain the prospective preprocessing algorithm and the details of the five different post-processing methods named: Method 1 + XGB, Method 2 + XGB, Method 3, Method 1 + Method 3, and Method 2 + Method 3. The algorithms are used to simultaneously increase *accuracy* and reliability as illustrated in Fig. 2.

*1) Preprocessing technique:* We suggest a Time-Series Augmentation (TSA) as a preprocessing technique before training the DNN and applying it on the training, validation, and testing sets. This technique inverts the time series sequence of each sample and generates a new augmented sample. For example, the RSSI and SINR, would both have the original sequence and the inverted one within an appropriate rolling window. During the preprocessing phase, we combine the original and inverted sequences of each variable to create four new samples. Generally, the technique could generate $2^{|Variables|}$ augmented samples from each variable that the DNN uses as an input. We use all of the original and the augmented samples to train our DNN as in Fig. 2. After the augmented version of each sample is processed by the DNN, the results are used as inputs in the post-processing algorithms. In Table I, we display an example of how to generate the four new augmented samples according to the preprocessing algorithm.

TABLE I: Output of the TSA.

|  | Sequence 1 | Sequence 2 |
|---|---|---|
| Sample 1 | Same | Same |
| Sample 2 | Same | Flipped |
| Sample 3 | Flipped | Same |
| Sample 4 | Flipped | Flipped |

*2) Post-processing Methods:* We explain the different postprocessing techniques in the system, specifically Methods 1, 2, and 3 (M1, M2, and M3). Each method utilizes the DNN outputs of all augmented samples generated in the preprocessing phase as explained in subsection II-C1. The final algorithm is formed by a combination of two of the techniques or by only a single one of them together with the XGB classifier.

*Method 1:* We apply this method on the probabilistic outputs of the DNN for all the augmented samples. Each output is in the one hot encoding form for binary classification. For example, $[\alpha, 1-\alpha]$ in which $\alpha$ is a number between zero and one. First, we define a filter variable $\beta$ and apply it on each DNN output. If the probabilities are in the range of $[\beta, 1-\beta]$, the values will be replaced by the constant 0.5. The purpose of this filtering is to neutralise the negative effect of low quality probabilities in the following steps. The parameter $\beta$ is a hyper parameter defined by the user and it is inversely proportional to the quality of the prediction. Thus, a smaller value of $\beta$ will lead to a higher number of samples being categorized as low quality prediction features.

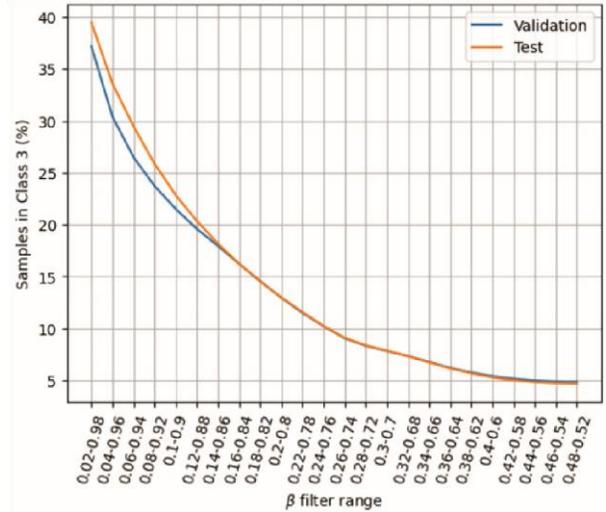

(a)

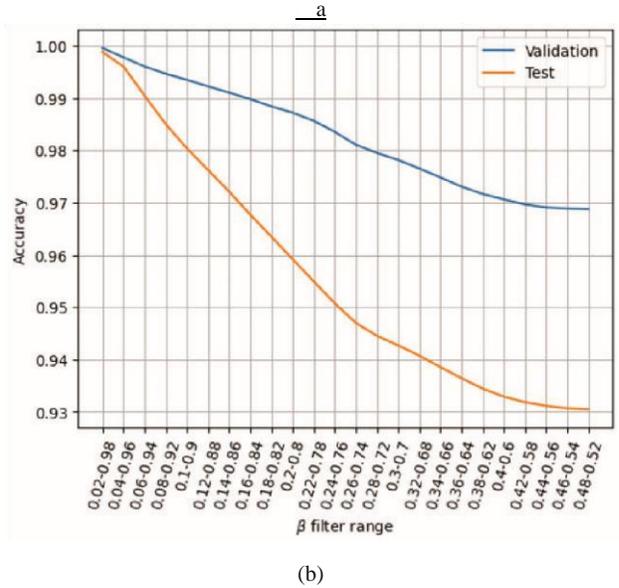

(b)

Fig. 3: Flexible ranges with a resolution of 0.02 between the *accuracy* of samples in classes 1 and 2 and the portion of class 3 samples based on the $\beta$ filter ranges: (a) Samples versus $\beta$ (b) Accuracy versus $\beta$.

Fig. 3a shows the effect of $\beta$ on the samples that are categorized in class 3 as percentages. For example, when $\beta$ is located in the 0.48-0.52 range the amount of features added in the third class is around 5% of the total features generated after the preprocessing phase, while 40% of the total samples are in class 3 if the $\beta$ range changes to 0.02-0.98. Fig. 3b analyzes the *accuracy* of the samples in classes 1 and 2 in the ranges specified by $\beta$. Here, the *accuracy* changes from 97% to 99.9% during the validation phase, and 93% to 99.88% during the test phase for same ranges mentioned in Fig. 3a. Both combined charts give insights into flexible *accuracy* based on the amount of samples allocated in class 3. After applying a filter, we calculate the average of augmented versions of a sample in a specific way. First, we calculate the summation of outputs and then round the result. As we have four outputs per sample, in our case, the expected values should be two integers in one hot encoding form (i.e., between 0 and 4). In general, the expected value after these processes should be between 0 and $N$; for $N$ augmented outputs. Then, we divide the results by $N$ to re-scale the output between zero and one and round the output to convert it to zero and one integers. If, after this process, the sample does not satisfy the one hot encoding rule, the sample will be rejected, and it is sent to class 3 where we use another algorithm to classify the rejected sample. Otherwise, the output is categorized as class 1 or class 2, as described in Fig. 2. We have two possibilities for features classified in class 3: Method 3 or XGB. The combination of Method 1 with Method 3 or XGB algorithms lead to the proposed algorithms M1+M3 or M1+XGB. The quality of the final result for combined methods depends on the quality of the ML algorithm as well as the filter $\beta$ adjustment.

*Method 2:* In Method 2, we convert outputs from a probabilistic to an integer form and apply a majority voting algorithm to them. If there is a tie after voting, for example, half of the outputs indicate that the features should be classified as class 1, and another half indicate they should be allocated to class 2, then the features are rejected and added in class 3 as low-scored features. Then, the ML algorithm XGB or Method 3 attempts to classify these low-scored features. Similar to Method 1, Method 2 is combined with the XGB algorithm or Method 3 to increase reliability and *accuracy*.

*Method 3:* Method 3 calculates the confidence of each output. The output with the maximum confidence value is selected as the final result. Our results suggest that combining Method 3 with the others, namely Methods 1 and 2, can satisfy several reliability metrics while boosting *accuracy*.

*Monte Carlo Technique:* To improve the reliability of DNN results, we recommend running only the DNN prediction phase several times and taking the probabilistic average of all runs before applying any of the methods. Some methods, such as Method 3 and its combinations, are heavily dependent on this technique and it directly impacts the quality of the final results. To minimize the latency imposed by the Monte Carlo approach, we aim to execute as few iterations as possible with satisfactory outcomes, such as 10 or 20 iterations. We use the DNN prediction average gain from 15 runs.

### D. Evaluation Metrics

We use well-known metrics proposed by [6] to measure the model's uncertainty, *accuracy*, and quality to compare method improvements with each other. These metrics are explained below:

Accuracy per Confidence. This metric is used in its visual form to analyze the calibration and uncertainty of the DNN model. The authors in [14, 15] refer to this chart as the reliability diagram. We calculate the metric by grouping samples based on their confidence values between interval ranges and estimating the *accuracy* of each group. We can define *accuracy* per confidence by looking at the chart of *accuracy* percentages for each group. In our deep network, we use one hot encoding output with softmax activation function and binary cross entropy loss function. As the studied DNN provides results in a one hot encoding probabilistic shape, we use the maximum probability value of predicted output classes as a confidence score and we grouped the values between 0.5 to 1 in interval ranges. The interval ranges are defined by the user which in our work we defined as 0.1. In a perfect world, the midpoint of each confidence interval would coincide with the *accuracy* at the same point. We illustrate the evaluation of the Accuracy per Confidence in Fig. 4. We do this by drawing ideal levels for the central values of each confidence interval.

Mean Confidence and Mean Accuracy. These two metrics are the total weighted average of confidence and *accuracy* for the number of samples per each confidence interval. These two numbers ought to be the same in a fair scenario regarding reliability and *accuracy*. However, these values typically exhibit skews to one extreme or the other in DNN architectures. Overconfidence occurs when the Mean Confidence is greater than the Mean Accuracy, while Under-confidence occurs when the second is greater than the first one. Calibrating uncertainty can help bring probabilistic outputs closer to the ideal levels with minimum or no loss in *accuracy* values.

Reliability Score. We define the difference between the MC and MA values by another metric which is denominated the Reliability Score. If the RS is equal to zero, the DNN achieves the optimal balance of *accuracy* and reliability. Overconfidence occurs when the Mean Confidence is greater than the Mean Accuracy, while Under-confidence occurs when the second condition is greater than the first. The authors in [6] demonstrate that a DNN with $N$ classes and $M$ inputs tends to be overconfident. Our study suggests that simple preprocessing and post-processing algorithms might change this classification.

Expected and Maximum Calibration Errors. At each confidence interval, the *accuracy* deviation away from the confidence interval center is considered as the error per each interval. The Expected Calibration Error is defined as the weighted error and the Maximum Calibration Error describes the maximum error per all intervals. In an ideal situation, these two errors would be zero [16].

Negative LogLikelihood Loss (NLL). This metric is known as cross-entropy loss and is used as a loss function for DNNs [17]. It is also utilized as a metric to measure the quality of the probabilistic model [18]. For each DNN output sample, we calculate the negative logarithm of the predicted probability of the ground truth class. After that, we normalize them and sum up all the outcomes. We show the final result in percentages. The ideal value of this metric is zero.

Brier Score Loss (BSL). This metric is defined by the square error of the predicted probability vector and ground truth values in one hot encoding form. This measure should be as close as possible to zero [6].

## III. EXPERIMENTAL RESULTS

In this section, we present the simulations results. We compare the results of the DNN using each of the five prospective methods to the results of the DNN with no method and the DNN with the XGB. We choose XGB because it is the best performing publicly available classifier applied to our dataset in terms of *accuracy* [9]. We use the Accuracy vs the Confidence Intervals Central Values and we evaluate the performance of each algorithm using the seven metrics previously mentioned in subsection II-D. To make the most accurate DNN predictions, we implement the Monte Carlo method across all algorithms.

Table II shows the metric details used to define the best algorithm performance. It is difficult to choose the best method based on only one metric or consider the best performance on each metric. There is no method that can satisfy the best performance on all metrics. Therefore, we highlight the top three results in each metric. After indicating the top three metric results in Table II, we notice that the combination of DNN and Methods 1 and 3 (Method 1+3) satisfies the most metrics. Comparison from No method to Method 1+3 shows an almost double increase in MCE, a considerable decrease in ECE, and minor differences in the remaining variables. Therefore, it is more necessary to lower the ECE, although the MCE error will increase. In second place, Method 3 can not only improve the total *accuracy*, but also satisfies most of the reliability metrics. Furthermore, this method achieves the closest to zero RS results followed closely by Method 1+3 compared to all the suggested algorithms.

The results of the M2+XGB indicates that the XGB can be used as a complementary algorithm to improve the *accuracy* of the DNN results (MA = 91.19 and ECE = 3.21). Even though the *accuracy* results of XGB algorithm alone was inferior (MA = 85.21). An comparison of M2+XGB with M1+XGB reveals that using class label outputs instead of probability values to calculate majority voting is recommended when we want to combine the XGB result with DNN.

The results of most of the combined algorithms placed the DNN in the over-confidence region (OC) except for the Method 1+3 algorithm and the DNN with No Methods applied. Both cases were in the under-confidence region (UC). The *accuracy* results of all the algorithms were similar except for the XGB and the Method 1+XGB. For example, the difference between the highest (M2 + XGB = 91.19) and the lowest values (No Method = 91.01) is 0.18%. However, the difference between both algorithms for the MCE and NLL indicators is 29.67 and 2.33, respectively. These differences highlight the *accuracy* discrepancies between the confidence interval values and decreases the reliability of the DNN. Therefore, it is fundamental to have DNN reliability evaluation results prior defining best performing architectures.

Fig. 4 shows the Accuracy versus the Confidence Intervals Central Values for the DNN combined with the five proposed

TABLE II: Key Performance Parameters for Reliability, Top three results for each metric are highlighted

| DNN+ | ECE (%) | MCE (%) | MC (%) | MA (%) | NLL | BSL (%) | OC or UC | RS = \|MA-MC\| (%) |
|---|---|---|---|---|---|---|---|---|
| No Method | 3.71 | 7.07 | 89.77 | 91.01 | 0.2 | 6.25 | UC | 1.24 |
| XGB | 7.22 | 27.77 | 92.43 | 85.21 | 0.63 | 12.45 | OC | 7.22 |
| Method 1 + XGB | 4.70 | 14.08 | 91.59 | 87.74 | 0.54 | 10.72 | OC | 3.85 |
| Method 2 + XGB | 3.21 | 36.74 | 94.41 | 91.19 | 2.53 | 8.32 | OC | 3.21 |
| Method 3 | 4.03 | 15.18 | 91.84 | 91.19 | 0.21 | 6.53 | OC | 0.65 |
| Method 1 + 3 | 2.19 | 12.37 | 90.26 | 91.19 | 0.22 | 6.82 | UC | 0.92 |
| Method 2 + 3 | 3.02 | 41.16 | 94.20 | 91.18 | 2.52 | 8.21 | OC | 3.02 |

ECE; Expected Calibration Error, MCE; Maximum Calibration Error, MC; Mean Confidence, MA; Mean Accuracy, NLL; Normalized Negative Log Likelihood, BSL; Brier Score Loss, OC; Over-Confidence, UC; Under-Confidence, RS; Reliability Score.

methods namely: Method 1 + XGB (M1+XGB), Method 2 +

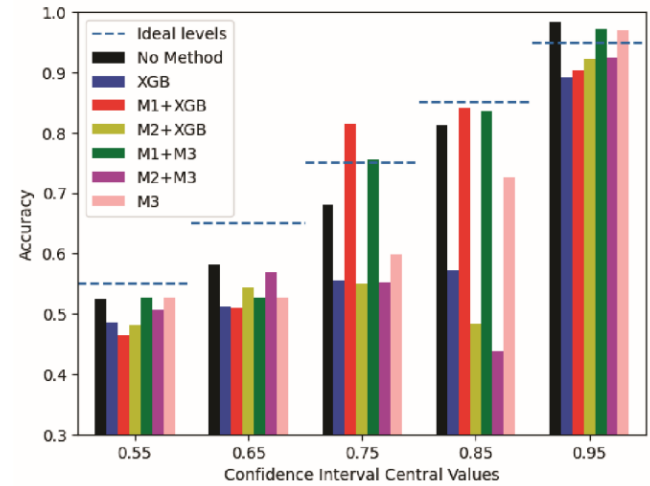

Fig. 4: The Accuracy versus the Confidence Values for the DNN added with M1+XGB, M2+XGB, M1+M3, M2+M3, M3, No Method, and XGB algorithm.

XGB (M2+XGB), Method 1 + Method 3 (M1+M3), Method 2 + Method 3 (M2+M3) and Method 3 (M3) added to the results related to the No Method and the XGB classifier. We also plotted the ideal levels that represent the best *accuracy* and reliability achievable by general DNNs. In the ideal scenario, all the algorithms should be as close as possible to this line, neither in the upper area nor in the lower area. The ideal levels divide the chart into two regions: the lower part represents over-confidence while the upper part symbolizes the under-confidence area. Ideally, the DNN should not be either too optimistic or too pessimistic about the data classification uncertainty.

Fig. 4 depicts that the DNN+M1+M3 algorithm consistently got closer to the ideal levels over the confidence intervals central values and only exceeds the ideal levels at 0.95 confidence central value. The second best algorithm is the DNN+M1+XGB. However this algorithm surpasses the ideal levels and make the DNN Network under-confident at the 0.75 confidence interval central value and it is the lowest performer at lower confidence values that ranges between 0.5 and 0.7. The DNN network shows rather good reliability at all confidence interval ranges as Fig. 4 illustrates when no methods were used and only Monte Carlo technique is applied. In general, all the algorithms struggle to increase *accuracy* at the lower ranges confidence intervals i.e., 0.5 and 0.7. In some cases, like with the M2+XGB, M2+M3 and XGB algorithms, the hard-to-raise *accuracy* extends to the 0.7 and 0.9 confidence value ranges. Also, most of the algorithms achieve a relatively good tradeoff between *accuracy* and reliability for the 0.95 confidence intervals central values. The XGB presented a reasonable performance in the confidence range of 0.9 and 1, but like other algorithms, it delivered a lower performance at the 0.5 and 0.9 range.

Based on the results shown in Fig. 4 and in table II, we propose using Method 1+3 to improve the *accuracy* and reliability of the DNN architecture. Also, the increase in the MCE is compensated by a reduction of the ECE.

## IV. CONCLUSION

It is expected to have ML mechanisms in 5G and 6G UAV communication systems. Therefore, it is fundamental to understand the uncertainties of the deep networks used in those systems and how reliable they are. In this study, we proposed five combined methods to increase *accuracy* and reliability concomitantly in binary classification deep networks applied to UAV security scenarios. By analyzing seven reliability metrics and the *accuracy* per confidence, Method 1 combined with Method 3 presented the best overall performance that satisfied most of the metrics by achieving the top three in each one. This algorithm reached an ECE of 2.19 and was closer to all ideal levels' values. Method 3 was the second-best performing algorithm in terms of reliability. With Method 2 + XGB, we showed that a lower performing ML algorithm can be combined with one of the proposed methods to increase the total DNN *accuracy*, but in terms of the reliability, this might not be a good option. By using proper preprocessing techniques (TSA) on time-series samples in terms of classification, we showed it was possible to generate different versions of each sample that provided diversity for post-processing techniques. Method 1 + Method 3, and Method 3 alone were good candidates for post-processing methods to calibrate uncertainty and increase the total *accuracy* of the DNN. Finally, four of the five methods presented were able to increase *accuracy*, but not all of them increased the reliability. As a result, network engineers and developers must take extra precaution when proposing DNN architectures and analyze them in terms of *accuracy* and reliability.

## ACKNOWLEDGMENT

This research received funding from the European Union's Horizon 2020 research and innovation programme under the Marie Sklodowska-Curie Project Number 813391. Also, this work was partially supported by Fundação para a Ciência e a Tecnologia and Instituto de Telecomunicações under Project UIDB/50008/2020.